\documentclass[10pt]{article}

\usepackage[letterpaper,margin=1in]{geometry}
\usepackage{amsmath,amssymb,booktabs,float,graphicx,microtype}
\usepackage{array,tabularx,multirow}
\usepackage{seqsplit}
\usepackage{natbib}
\usepackage{xcolor}
\usepackage{hyperref}
\usepackage{url}
\usepackage{fancyhdr}
\usepackage{tikz}
\usetikzlibrary{arrows.meta,positioning}

\definecolor{mgtblue}{RGB}{30,101,170}
\definecolor{mgtorange}{RGB}{220,120,0}
\hypersetup{colorlinks=true,citecolor=mgtblue,linkcolor=mgtblue,urlcolor=mgtblue}
\newcommand{\E}{\mathbb{E}}
\newcommand{\Prob}{\mathbb{P}}
\newcommand{\R}{\mathbb{R}}
\newcommand{\method}{MGT-B}
\newcommand{\pp}{\,pp}
\newcommand{\hashid}[1]{\texttt{\seqsplit{#1}}}
\newcommand{\authorblock}[4]{%
  \noindent
  \begin{minipage}[t]{0.76\textwidth}
    \textbf{#1}\\[1pt]
    \textit{#2}\\[1pt]
    \textit{ORCID: \url{#3}}
  \end{minipage}%
  \hfill
  \begin{minipage}[t]{0.20\textwidth}
    \raggedleft\texttt{#4}
  \end{minipage}
  \par\vspace{3mm}
}

\title{\textbf{CUSUM-Shaped Inference-Time Monitoring and Targeted Re-Decoding for Quantized Small Language Model Reasoning}}
\author{}
\date{}

\begin{document}
\maketitle
\vspace{-10mm}

\authorblock{El Hassane Ettifouri}{Novelis Research, Paris, France}{https://orcid.org/0000-0001-5299-9053}{eettifouri@novelis.io}
\authorblock{Ayoub Belfatmi}{Novelis Research, Paris, France}{https://orcid.org/0009-0005-4010-794X}{abelfatmi@novelis.io}
\authorblock{Mahaman Sanoussi Yahaya Alassan}{Novelis Research, Paris, France}{https://orcid.org/0009-0006-0825-4701}{syahaya@novelis.io}
\authorblock{Walid Dahhane}{Novelis Research, Paris, France}{https://orcid.org/0000-0001-5387-3380}{wdahhane@novelis.io}
\vspace{4mm}

\begin{abstract}
Quantized small reasoning models can enter repetitive or otherwise unproductive trajectories, yet standard decoding does not adapt to the trajectory as it unfolds. We study \method{}, a fixed, weight-preserving controller that converts overlapping windows of uncertainty, repetition, and local-change features into position-conditional empirical tail probabilities. It accumulates mixture betting factors with a CUSUM-shaped reset, and, after an alarm, restores a coherent earlier token and key--value-cache state before constrained re-decoding. On MATH-500, a paired three-seed evaluation over 1,500 generations per method raises exact-normalized accuracy from 54.73\% for vanilla decoding to 56.40\% (+1.67 percentage points; problem-clustered bootstrap 95\% CI $[+0.47,+2.80]$), while a prospectively profiled random-intervention control reaches 54.60\%. The gain is positive in all three seeds and costs 5.14\% more sampled tokens. Seed-0 ablations show that rollback alone does not explain the result and that an isolated repetition penalty is harmful. Five-sample self-consistency reaches 70.0\% but uses about $4.84\times$ as many tokens as \method{}. On the harder, non-overlapping Omni-MATH evaluation, however, \method{} obtains 16.60\% versus 16.67\% for vanilla ($-0.07$ points; clustered 95\% CI $[-0.33,+0.20]$) with 2.10\% more sampled tokens. Thus, \method{} provides a modest, reproducible local improvement on MATH-500 in the studied configuration, but the effect does not transfer to Omni-MATH and should not be interpreted as a general improvement in mathematical reasoning.
\end{abstract}

\section{Introduction}

Distilled reasoning models make long chain-of-thought generation feasible on limited hardware, while low-bit post-training quantization further reduces deployment cost. However, quantization can be accompanied by sharper losses in mathematical reasoning quality, longer trajectories, and early errors that propagate through later steps \citep{li2025quantization,liu2025quantization,lotfi2026quantized}. These observations motivate an intervention point inside decoding: if an unproductive trajectory can be recognized while it is unfolding, compute can be redirected before the model exhausts its budget or commits to an unrecoverable suffix.

Autoregressive reasoning is path dependent. Repetitive degeneration is well documented in neural generation \citep{holtzman2020curious}, and recent analyses identify circular, self-reinforcing loops as a distinct failure mode of long reasoning trajectories \citep{duan2026circular}. The relevant question is therefore not whether one token is individually surprising, but whether a recent segment provides accumulating evidence of drift and whether a targeted intervention can still recover it.

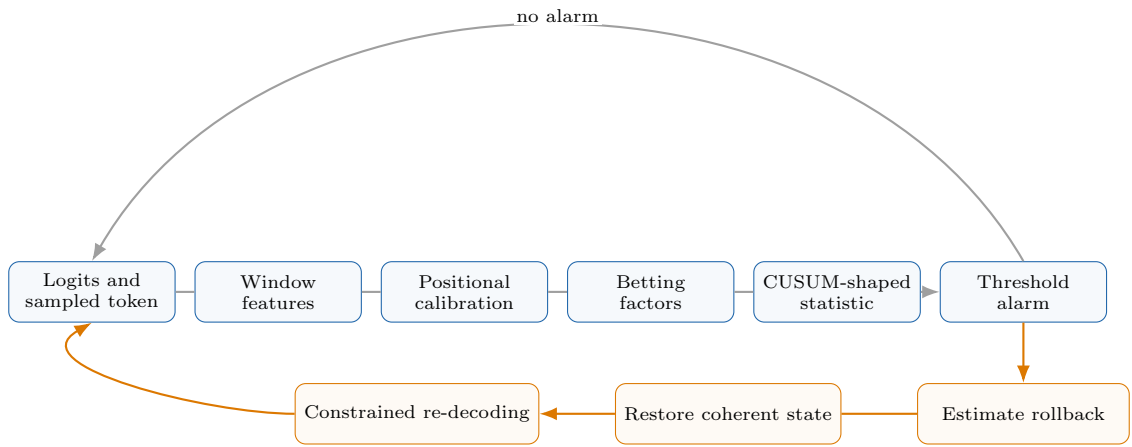
\begin{figure}[t]
\centering
\begin{tikzpicture}[font=\scriptsize,>=Latex,node distance=2.5mm and 2.5mm,
  proc/.style={draw=mgtblue,rounded corners,minimum height=8mm,minimum width=22mm,align=center,fill=mgtblue!4},
  repair/.style={draw=mgtorange,rounded corners,minimum height=8mm,minimum width=28mm,align=center,fill=mgtorange!4},
  flow/.style={->,thick,draw=gray!75}, alarm/.style={->,thick,draw=mgtorange}]
\node[proc] (logits) {Logits and\\sampled token};
\node[proc,right=of logits] (features) {Window\\features};
\node[proc,right=of features] (cal) {Positional\\calibration};
\node[proc,right=of cal] (bet) {Betting\\factors};
\node[proc,right=of bet] (cusum) {CUSUM-shaped\\statistic};
\node[proc,right=of cusum] (thresh) {Threshold\\alarm};
\node[repair,below=8mm of thresh] (rollback) {Estimate rollback};
\node[repair,left=10mm of rollback] (restore) {Restore coherent state};
\node[repair,left=10mm of restore] (redecode) {Constrained re-decoding};
\draw[flow] (logits)--(features)--(cal)--(bet)--(cusum)--(thresh);
\draw[alarm] (thresh)--(rollback);
\draw[alarm] (rollback)--(restore)--(redecode);
\draw[alarm] (redecode.west) to[out=180,in=210] (logits.south);
\draw[flow] (thresh.north) to[out=120,in=60] node[above,fill=white,inner sep=1pt]{no alarm} (logits.north);
\end{tikzpicture}
\caption{Monitoring-and-control loop. An alarm activates rollback and state-consistent re-decoding; without an alarm, decoding continues unchanged. An alarm is an operational trigger, not a certified semantic error label.}
\label{fig:controller}
\end{figure}

We study a fixed external controller that observes logits and sampled tokens but never changes model weights. It combines uncertainty with explicit repetition and local entropy-change features, calibrates the resulting window score by generation position, and accumulates a CUSUM-shaped statistic. At an alarm, it estimates a rollback point, restores all coupled state, and re-decodes locally under modified sampling constraints. Figure~\ref{fig:controller} summarizes the implemented loop.

The statistical terminology requires care. Classical CUSUM originates in sequential change detection \citep{page1954continuous}. E-processes and e-detectors provide time-uniform guarantees only under conditional validity assumptions \citep{shafer2021testing,ramdas2023gametheoretic,shin2024edetectors}. Here, empirical probabilities are computed from dependent, overlapping language-model windows. We therefore call the construction \emph{CUSUM-shaped}; its threshold is an empirical operating point, not a theoretical false-alarm certificate.

This paper makes three empirical contributions:
\begin{enumerate}
  \item a window-based, positionally calibrated monitor coupled to state-consistent targeted re-decoding;
  \item a paired MATH-500 evaluation on the same 500 problems under three generation seeds, including a prospectively profiled random-intervention control and explicit sampled-token accounting;
  \item seed-0 component ablations, a five-sample self-consistency baseline, and a three-seed external evaluation on a non-overlapping Omni-MATH test set.
\end{enumerate}

\method{} improves the studied quantized 1.5B model modestly and reproducibly on MATH-500, but it neither matches the unconstrained accuracy of self-consistency nor transfers its gain to Omni-MATH.

\section{Related Work}

\paragraph{Online signals for reasoning.}
Entropy and confidence have been used to diagnose or control reasoning trajectories. Reasoning-path deviation monitoring uses local entropy behavior to detect overthinking online \citep{guan2026deviation}, while other work connects entropy dynamics to underthinking and reasoning quality. Low entropy is ambiguous, however: it may reflect confident progress or a confident loop. \method{} therefore combines uncertainty with direct repetition and confidence-change signals.

\paragraph{Degeneration and circular reasoning.}
Neural text degeneration can be self-reinforcing \citep{holtzman2020curious}. In long reasoning models, semantic repetition can precede overt lexical loops; \citet{duan2026circular} use CUSUM for early loop prediction. Recent work also tracks convergence followed by semantic oscillation \citep{wei2026evolution}, while PUMA detects redundant reasoning steps with a lightweight semantic encoder and verifies whether early exit is safe \citep{min2026puma}. These methods are direct prior art for semantic monitoring. The present controller instead combines token-level signals with positional empirical calibration and uses an alarm to trigger coherent rollback and re-decoding rather than only terminating the trace. Neither CUSUM itself nor semantic repetition detection is claimed as novel.

\paragraph{Test-time scaling and revision.}
Self-consistency samples independent paths and selects by agreement \citep{wang2023selfconsistency}; adaptive test-time scaling allocates compute according to difficulty \citep{snell2024scaling}. Tree of Thoughts searches among explicit states \citep{yao2023tree}, Self-Refine iterates feedback and revision \citep{madaan2023selfrefine}, and self-backtracking methods learn or prompt a decision to reverse course \citep{yang2025selfbacktracking}. Stochastic backtracking can also retain a pool of historical prefixes and use a process reward model to revisit them \citep{tran2026frontier}; bursts of late backtracking have separately been identified as markers of overthinking \citep{rezazadeh2026shape}. \method{} instead follows one trajectory until an external, training-free sequential signal activates, then restores one earlier token and cache state. It maintains neither a search tree nor a prefix pool and uses no process reward model. Its relevant comparison is therefore accuracy at explicit sampled-token cost.

\paragraph{Process supervision and sequential monitoring.}
Process supervision trains verifiers on intermediate steps \citep{lightman2024verify}. \method{} uses neither step labels nor a learned verifier: it detects deviation relative to an empirical reference, not semantic incorrectness. Likewise, its CUSUM-shaped recurrence is operationally inspired by sequential testing but does not satisfy a proved e-process condition.

\section{Positioning and Contribution}

The contribution of \method{} is integrative rather than the invention of any one primitive. Entropy, token confidence, repetition, CUSUM, rollback, and constrained decoding all have prior uses. The specific construction studied here combines six signals over overlapping windows and conditions their empirical calibration on generation position. It accumulates a mixture of betting factors with a CUSUM-shaped reset. At an alarm, it estimates a rollback point from the reset history and restores tokens, KV cache, feature state, n-gram state, and detector state jointly before re-decoding. Such a coupling is operationally significant: an inconsistent rollback can invalidate both the cache and the monitor, while a detector with a flawed repair policy cannot improve final answers.

An earlier token-level controller used a global healthy baseline, a linear betting process, and a simpler intervention policy \citep{anonymous2026prior}. The current version changes the monitoring unit, positional calibration, betting construction, rollback localization, state restoration, and  adds marker-free re-decoding. The experiments therefore evaluate the complete revised controller. They do not yet isolate the benefit of each architectural change relative to the earlier controller, which remains an explicit comparison for future work.

\method{} is not a general state-of-the-art reasoning method. Its empirical contribution is a selective, low-overhead improvement on one MATH-500 configuration, accompanied by evidence that the same policy does not transfer automatically to Omni-MATH.

\section{Problem Setup}

Let $x$ be a prompt and $W_t\in\mathcal{V}$ the token sampled at step $t$. Before sampling, the model produces
\begin{equation}
  p_t(v)=\Prob(W_t=v\mid\mathcal{F}_{t-1}), \qquad v\in\mathcal{V},
\end{equation}
where $\mathcal{F}_{t-1}$ contains the retained token prefix, model cache, monitor state, and decoding decisions. For the $j$th token window $B_j$, the monitor produces $Z_j\in\R^6$, a scalar score $s_j$, a calibrated upper-tail probability $q_j$, a betting factor $e_j$, and a reset statistic $S_j$. The alarm time is
\begin{equation}
  \tau=\inf\{j\geq1:S_j\geq h\}.
\end{equation}

We distinguish \emph{sampled tokens}, which count every decoding event including abandoned branches, from \emph{emitted tokens}, which remain in the final completion. Deleted tokens are sampled and later removed by rollback. The primary endpoint is final-answer accuracy; secondary endpoints include corrections, regressions, interventions, truncations, extraction, and token cost.

\section{Method}

\subsection{Why centered token log-probability is insufficient}

Let $\ell_t=\log p_t(W_t)$ and $H_t=-\sum_v p_t(v)\log p_t(v)$. Since $W_t$ is sampled from $p_t$,
\begin{equation}
  \E[\ell_t\mid\mathcal{F}_{t-1}]=-H_t,
  \qquad D_t=\ell_t+H_t,
  \qquad \E[D_t\mid\mathcal{F}_{t-1}]=0.
\end{equation}
It is a valid sampling-consistency statement, but not a health signal. In confident repetition, $p_t(W_t)\approx1$, so both $\ell_t$ and $H_t$ approach zero and the centered increment may remain silent. The monitor therefore uses chosen-token log-probability only as one component of a multifeature window score.

\subsection{Window score and positional calibration}

For each 64-token window with stride 32, the controller averages entropy and chosen-token log-probability, measures repeated generated 6--8-grams, records increased confidence on repeated n-grams, and computes positive and negative local entropy shifts. The score is
\begin{equation}
  s_j=w_H\bar H_j+w_\ell(-\bar\ell_j)+w_RR_j+w_DD_j+w_+L_j^+ +w_-L_j^- ,
\end{equation}
with fixed weights $(0.15,0.10,0.20,0.35,0.18,0.02)$.

Let $\mathcal{C}_{b(j)}$ be the reference scores whose window endpoints share position bucket $b(j)$. The empirical upper-tail probability is
\begin{equation}
  q_j=\frac{1+\sum_{u\in\mathcal{C}_{b(j)}}\mathbf{1}\{u\geq s_j\}}
  {|\mathcal{C}_{b(j)}|+1}.
\end{equation}
This corrects coarse positional drift but does not render overlapping windows conditionally exchangeable.

\subsection{CUSUM-shaped accumulation and intervention}

For beta exponents $\Gamma=\{0.1,0.3,0.5,0.7\}$, the incremental factor is
\begin{equation}
  e_j=\frac{1}{|\Gamma|}\sum_{\gamma\in\Gamma}\gamma\,\widetilde q_j^{\gamma-1},
  \qquad \widetilde q_j=\min(1,\max(10^{-6},q_j)).
\end{equation}
The implemented reset statistic is
\begin{equation}
  S_j=\max(0,S_{j-1})+\log e_j,\qquad S_0=0.
\end{equation}
The condition $\E[e_j\mid\mathcal{F}_{j-1}]\leq1$ is not proved, so $e_j$ is called a betting factor rather than a validated incremental e-value.

At an alarm, the controller finds the most recent reset region, maps it to a token position, extends rollback by a 64-token margin, and truncates tokens, KV cache, window features, n-gram state, and detector state coherently. It then re-decodes at temperature 0.6 with repetition penalty 1.1, suspect n-gram blocking, two refractory windows, and at most three rerolls. No correction marker is injected. When no alarm occurs, the controlled output remains token-identical to vanilla under paired randomness.

\section{Experimental Setup}

\subsection{Model and generation}

All experiments use \path{deepseek-ai/DeepSeek-R1-Distill-Qwen-1.5B} at revision \hashid{ad9f0ae0864d7fbcd1cd905e3c6c5b069cc8b562}. The model is loaded in bitsandbytes INT4 FP4 without double quantization and with FP16 computation on an NVIDIA RTX A5000. The prompt requests step-by-step reasoning and a final \texttt{\#\#\#\# <answer>}; generation is capped at 20,000 new tokens. Item-level seeds are paired across methods.

MATH-500 uses all 500 benchmark problems and deterministic exact-normalized answer matching. The scorer normalizes common LaTeX, fraction, root, spacing, and decimal forms but is not a general symbolic-equivalence checker. The main comparison evaluates seeds 0, 1, and 2, for 1,500 paired units per method.

Omni-MATH is constructed from \texttt{KbsdJames/Omni-MATH} after NFKC normalization, SHA-256 deduplication, and removal of any prompt overlapping MATH-500. A deterministic, stratified split fixes 300 reference, 300 development, and 500 test problems before generation. Test problems are evaluated under the same three seeds.

\subsection{Calibration and controls}

For MATH-500, the positionally calibrated full monitor uses $h=11.4293$, with 3 alarms among 62 eligible healthy development trajectories (4.84\%). For Omni-MATH, the feature distributions and threshold are recalibrated without correctness labels from the test set; $h=47.6874$ yields 14 alarms among 299 eligible development trajectories (4.68\%). All score weights, window settings, CUSUM recurrence, and repair parameters remain unchanged.

The principal controls are vanilla decoding and \texttt{matched\_random}. The latter draws intervention frequency, position, and rollback length from a development profile produced by \method{} without test labels, then applies the same repair. Seed-0 MATH-500 ablations evaluate repetition-only and non-reset monitors; fixed rollback; rollback-only; rollback combined separately with lower temperature, repetition penalty, or n-gram blocking; and five-sample self-consistency.

\subsection{Scoring and statistical analysis}

Omni-MATH requires broader equivalence judgments. Exact normalized matches and simple numeric contradictions are resolved deterministically; remaining cases are blind-scored by \texttt{gemini-3.5-flash-lite} at temperature zero with structured outputs and \texttt{ABSTAIN} counted as incorrect. Payloads omit method, seed, prior verdicts, and aggregate metrics. Grouped and individual judgments agree on 94\% of 50 audited pairs. Because absolute Omni-MATH accuracy differs substantially under an alternative local judge, we report judge sensitivity explicitly.

Accuracy differences are paired at the problem--seed level. The main 95\% intervals use 10,000 bootstrap samples clustered by problem, so all three seeds of a problem remain together. Exact McNemar tests at the unit level are descriptive; declared families of comparisons against vanilla use Holm correction. Seed-0 ablations and category breakdowns are exploratory.

\section{Results}

\subsection{MATH-500 main comparison}

Table~\ref{tab:math-main} reports the three-seed comparison. \method{} improves on vanilla for every seed and reaches 846/1,500 correct (56.40\%) versus 821/1,500 (54.73\%). The paired gain is +1.67 points with a problem-clustered 95\% interval of $[+0.47,+2.80]$. It produces 53 corrections and 28 regressions; exact McNemar $p=0.0073$ and Holm-adjusted $p=0.0146$. \texttt{Matched\_random} remains indistinguishable from vanilla and is 1.80 points below \method{}.

\begin{table}[H]
\centering
\caption{MATH-500 accuracy on the same 500 problems under three paired generation seeds.}
\label{tab:math-main}
\small
\begin{tabular}{lrrrrr}
\toprule
Seed & Vanilla & \method{} & Random & MGT-B $-$ vanilla & MGT-B $-$ random\\
\midrule
0 & 55.2\% & 56.2\% & 55.0\% & +1.0\pp & +1.2\pp\\
1 & 54.0\% & 56.8\% & 54.0\% & +2.8\pp & +2.8\pp\\
2 & 55.0\% & 56.2\% & 54.8\% & +1.2\pp & +1.4\pp\\
\midrule
Aggregate & 821/1500 & 846/1500 & 819/1500 & +1.67\pp & +1.80\pp\\
 & 54.73\% & 56.40\% & 54.60\% & $[+0.47,+2.80]$ & $[+0.47,+3.13]$\\
\bottomrule
\end{tabular}
\end{table}

The random control intervenes on 9.4\% of units versus 22.6\% for \method{}. The development-derived profile therefore does not equalize activation frequency ex post, thus limiting component attribution. Yet the random policy's $-0.13$-point difference from vanilla shows that additional interventions alone do not reproduce the observed gain.

\subsection{MATH-500 compute and selectivity}

\begin{table}[H]
\centering
\caption{Generation behavior for the MATH-500 main comparison. Token values are per unit.}
\label{tab:math-cost}
\small
\begin{tabular}{lrrrrr}
\toprule
Method & Intervention & Sampled tokens & Cost vs. vanilla & Truncated & Extraction\\
\midrule
Vanilla & 0.0\% & 4,366.9 & --- & 0.13\% & 99.73\%\\
\method{} & 22.6\% & 4,591.3 & +5.14\% & 1.20\% & 99.80\%\\
Matched random & 9.4\% & 4,360.7 & $-0.14$\% & 0.27\% & 99.73\%\\
\bottomrule
\end{tabular}
\end{table}

Across 1,500 units, \method{} samples 336,543 more tokens than vanilla and deletes 317,312 tokens through rollback. Its retained output length rises by only 0.29\%, showing why emitted length understates decoding cost. All 1,161 non-alarmed \method{} trajectories are token-identical to vanilla. Truncation rises from 2/1,500 to 18/1,500, an operational cost not captured by mean accuracy.

\subsection{Seed-0 ablations}

Table~\ref{tab:ablations} separates monitor and repair variants on the same 500 seed-0 problems. Fixed rollback is numerically best among the single-path controllers at 56.8\%, but is not distinguishable from full \method{} at 56.2\%. Rollback alone reaches 54.0\%, indicating that erasing a suffix without changing continuation policy is insufficient. An isolated repetition penalty is harmful: 52.2\%, four points below full \method{}; the contrast remains significant after Holm correction. N-gram blocking alone remains close to vanilla. Repetition-only monitoring is parsimonious (35 alarms) and reaches 56.0\%, but its paired effect rests on only four favorable changes.

\begin{table}[H]
\centering
\caption{Exploratory MATH-500 seed-0 ablations. Sampled tokens are per problem.}
\label{tab:ablations}
\scriptsize
\begin{tabular}{lrrrrr}
\toprule
Method & Correct & Accuracy & vs. vanilla & Alarmed & Sampled tokens\\
\midrule
Vanilla & 276 & 55.2\% & --- & 0 & 4,332.5\\
Full \method{} & 281 & 56.2\% & +1.0\pp & 107 & 4,519.4\\
Matched random & 275 & 55.0\% & $-0.2$\pp & 55 & 4,270.5\\
Repetition-only monitor & 280 & 56.0\% & +0.8\pp & 35 & 4,321.4\\
Monitor without reset & 281 & 56.2\% & +1.0\pp & 79 & 4,276.5\\
Fixed 1,024-token rollback & 284 & 56.8\% & +1.6\pp & 107 & 4,552.1\\
Rollback only & 270 & 54.0\% & $-1.2$\pp & 107 & 4,629.0\\
Rollback + temperature & 274 & 54.8\% & $-0.4$\pp & 107 & 5,032.5\\
Rollback + repetition penalty & 261 & 52.2\% & $-3.0$\pp & 107 & 4,743.3\\
Rollback + n-gram blocking & 277 & 55.4\% & +0.2\pp & 107 & 4,665.4\\
Self-consistency (5) & 350 & 70.0\% & +14.8\pp & 0 & 21,867.7\\
\bottomrule
\end{tabular}
\end{table}

Five-sample self-consistency provides a much higher accuracy ceiling, 70.0\%, but consumes 10,933,855 total sampled tokens, approximately $4.84\times$ the cost of full \method{} on this seed. At least one of its five candidates is correct for 79.0\% of problems, leaving a further 9-point gap attributable to candidate selection.

\subsection{Omni-MATH external evaluation}

No accuracy gain transfers to Omni-MATH (Table~\ref{tab:omni}). Vanilla obtains 250/1,500 correct (16.67\%), full \method{} 249/1,500 (16.60\%), and \texttt{matched\_random} 254/1,500 (16.93\%). Only five \method{}--vanilla verdicts differ: two corrections and three regressions. The clustered 95\% interval for the $-0.07$-point contrast is $[-0.33,+0.20]$, excluding neither a very small benefit nor a very small harm.

\begin{table}[H]
\centering
\caption{Omni-MATH accuracy and paired comparisons under the final blind scoring pipeline.}
\label{tab:omni}
\small
\begin{tabular}{lrrrr}
\toprule
Method & Correct & Accuracy & Difference vs. vanilla & Clustered 95\% CI\\
\midrule
Vanilla & 250/1500 & 16.67\% & --- & ---\\
\method{} & 249/1500 & 16.60\% & $-0.07$\pp & $[-0.33,+0.20]$\\
Matched random & 254/1500 & 16.93\% & +0.27\pp & $[-0.07,+0.67]$\\
\bottomrule
\end{tabular}
\end{table}

The per-seed \method{}--vanilla differences are 0.0, $-0.2$, and 0.0 points. \method{} activates on only 69/1,500 units (4.60\%), compared with 22.6\% on MATH-500, and samples 7,027.7 tokens per unit versus 6,883.0 for vanilla (+2.10\%). It deletes 185,184 tokens and truncates 8 units, versus 5 for vanilla. Thus, it does not improve the Omni-MATH accuracy--token tradeoff.

Absolute Omni-MATH accuracy is judge-sensitive: the alternative local Omni-Judge pipeline agrees with the final pipeline on 70.47\% of 3,332 comparable units and places vanilla and \method{} at 35.60\% and 35.53\%, respectively. The relative difference remains $-0.07$ points under both judges. The evidence for a null relative effect is therefore directionally robust, whereas the absolute benchmark level is not.

\section{Discussion}

The MATH-500 results support a selective monitoring-and-repair effect. The full controller changes only alarmed trajectories, produces 25 more corrections than regressions over three seeds, outperforms a development-profiled random intervention control, and does so with a 5.14\% sampled-token premium. It is a modest efficiency gain rather than a replacement for broader test-time scaling.

Rollback alone is not enough: it reaches only 54.0\%, and neither lower temperature nor repetition penalty in isolation reconstructs the full effect. The repetition penalty alone is actively harmful in the tested repair. Conversely, fixed rollback is not measurably worse than adaptive rollback, and removing the reset changes individual outcomes without changing mean accuracy. Taken together, these ablations show the intervention is composite and fragile. The evidence validates the complete MATH-500 policy more strongly than any particular feature or rollback-localization rule.

After recalibration for Omni-MATH, the threshold rises from 11.43 to 47.69 and the intervention rate falls from 22.6\% to 4.6\%. One plausible interpretation is that Omni-MATH errors are dominated by missing capability or deeper reasoning failures rather than the localized repetition and drift modes targeted by the monitor. The experiment does not causally separate monitor false negatives, conservative calibration, and ineffective repair, so this explanation remains a hypothesis.

\section{Limitations}

The study evaluates one 1.5B distilled model, one INT4 quantization configuration, and mathematical reasoning only. Three generation seeds quantify sampling variability on the same 500 problems but do not create 1,500 independent tasks; that is why inference clusters by problem. Seed-0 ablations are exploratory and cannot establish small component effects.

The empirical tail probabilities are not proved conditionally super-uniform, so the statistic is neither a validated e-process nor an e-detector. Correct, non-truncated reference trajectories are only outcome-based proxies for locally healthy reasoning. No temporal error labels or true changepoints are available, preventing estimates of detector precision, recall, or detection delay.

MATH-500 scoring is deterministic but not a general symbolic equivalence checker. Omni-MATH relies partly on a learned judge; although the relative \method{}--vanilla difference is identical under two pipelines, absolute accuracy is judge-sensitive. The matched-random intervention rate is lower than \method{}'s ex post rate on both benchmarks, so it is informative but not perfectly exposure-matched.

Sampled tokens measure decoding work more faithfully than emitted tokens, but wall-clock latency, energy, peak memory, and hardware variability are not controlled systems endpoints. \method{} also raises truncation frequency on both benchmarks. Finally, the shift in calibration threshold and activation frequency across datasets shows that transfer cannot be assumed even when controller weights and repair rules are fixed.

\section{Future Work}

The next experiments should prioritize explanation and mechanism improvement rather than adding undirected variants. In particular, the Omni-MATH result creates a useful diagnostic setting: it separates a dataset on which the complete policy helps from one on which the same feature weights and repair rules do not.

\subsection{Diagnosing the transfer failure}

The first objective is to determine whether the Omni-MATH failure comes from the monitor, calibration, repair, or base-model capability. Doing so requires comparing the distributions of $s_j$, $q_j$, $e_j$, and $S_j$ across datasets as functions of position and problem difficulty, rather than comparing only final alarm rates. A stratified sample of incorrect non-alarmed Omni-MATH trajectories can then be annotated using a predeclared taxonomy: initial strategy error, execution error, lexical loop, paraphrastic repetition, oscillation between strategies, semantic stagnation, overthinking, and non-termination. These annotations would reveal whether useful alarms are suppressed by the higher threshold or whether the dominant errors lie outside the monitor's target class.

Monitor quality and repair quality also have to be separated by replay. Given one fixed set of alarm times and rollback positions, multiple repair policies can be evaluated on exactly the same triggered units; conversely, alternative monitors can share one fixed repair. A difficulty-conditioned threshold or intervention policy is then a natural extension, but its conditioning rule and all cutoffs must be selected entirely on development data before evaluation.

\subsection{Isolating monitor and repair components}

A compact confirmatory component matrix should repeat the most informative seed-0 comparisons over three seeds. The monitor tests are: full versus repetition-only features, reset versus non-reset accumulation under separately calibrated but equally targeted development alarm rates, and the current window controller versus the earlier token-level controller on identical items and seeds. The rollback test must compare adaptive and fixed rollback while holding the alarm time, repair, and per-item token budget constant.

A factorial evaluation of the repair covers lower temperature, repetition penalty, and suspect n-gram blocking. The current evidence makes removal of the repetition penalty a particularly important candidate: used alone, it is significantly harmful, but its interaction with the other repair components is not identified. The primary random control has to remain prospectively profiled; a secondary exposure-matched analysis should additionally enforce the same number of interventions as \method{}. All comparisons must report corrections, regressions, sampled tokens, deleted tokens, and truncations, not accuracy alone.

\subsection{A multi-scale semantic monitor}

The present monitor is strongest at explicit lexical reuse and local uncertainty change. It can miss paraphrases, slow loss of novelty, or oscillations between semantically similar strategies. A promising extension is therefore a periodic, multi-scale semantic channel computed over reasoning segments. Candidate features include nearest-prior-segment similarity, the slope of semantic novelty, return to an earlier semantic state, two-state oscillation, and persistence without measurable progress. Together, these features could distinguish at least four failure modes: lexical looping, paraphrastic repetition, strategy oscillation, and semantic stagnation.

Semantic monitoring is not new: convergence and redundancy signals have already been used for loop prediction and early exit \citep{duan2026circular,wei2026evolution,min2026puma}. The open contribution is to integrate such signals into positional calibration and the existing betting mixture, use the inferred semantic change region to localize rollback, and then test whether re-decoding can recover rather than merely stop the trace. To preserve the intended local-deployment regime, the semantic representation must be lightweight: for example, a small projection of hidden states or a compact encoder evaluated only every few windows. Fixed additive fusion represents the first baseline; any learned gate conditioned on position or estimated difficulty should be regularized, frozen on development data, and compared against the fixed mixture to quantify overfitting.

\subsection{Sequential validity, systems efficiency, and escalation}

Synthetic or human-annotated changepoints would permit direct measurement of detection precision, recall, and delay. They would also support testing a restart-valid e-detector or a conditionally calibrated alternative. Such work must keep operational performance and statistical validity distinct: a useful controller need not have an anytime-valid guarantee, and a valid detector need not identify semantic errors unless its null and labels are appropriate.

Implementation work can reduce monitoring cost independently of accuracy. The entropy and chosen-token log-probability calculations can share one log-softmax pass and n-gram occurrences can be indexed incrementally rather than searched across the retained trajectory. Semantic features can be cached at segment boundaries. 

Finally, \method{} can serve as the first stage of an adaptive compute policy. A low-cost intervention may be attempted once; persistent uncertainty, repeated alarms, or a high predicted difficulty can then escalate to self-consistency or a larger model. Evaluating this cascade across several budgets would place the controller on an accuracy--cost frontier and test whether selective repair and multi-sample scaling are complementary rather than competing alternatives.

\section{Conclusion}

\method{} combines positionally calibrated window monitoring, CUSUM-shaped accumulation, coherent rollback, and constrained re-decoding without changing model weights. On DeepSeek-R1-Distill-Qwen-1.5B in INT4, it improves MATH-500 from 54.73\% to 56.40\% across three paired seeds, with a clustered confidence interval entirely above zero and a 5.14\% sampled-token premium. Seed-0 ablations indicate that rollback alone does not explain the gain and that repair components cannot be assumed beneficial in isolation.

The gain does not transfer to Omni-MATH: accuracy changes from 16.67\% to 16.60\% while sampled-token cost rises by 2.10\%. The defensible claim is therefore that targeted inference-time monitoring can yield a small, cost-efficient local improvement when its drift signals align with a benchmark's failure modes. It is not yet a general method for improving mathematical reasoning. The most informative next step is to explain the transfer failure, then test a multi-scale semantic monitor and a better-isolated repair under prospectively fixed, multi-seed protocols. Broader model and quantization coverage should follow only after those mechanisms are understood.

\appendix

\section{Controller Configuration}

\begin{table}[H]
\centering
\caption{Fixed controller parameters and dataset-specific calibrated thresholds.}
\small
\begin{tabular}{ll}
\toprule
Component & Value\\
\midrule
Window length / stride & 64 / 32\\
Generated n-gram lengths & 6--8\\
Score weights $(H,-\ell,R,D,L^+,L^-)$ & $(0.15,0.10,0.20,0.35,0.18,0.02)$\\
Betting exponents & $(0.1,0.3,0.5,0.7)$\\
Empirical probability clip & $10^{-6}$\\
MATH-500 threshold & 11.4293\\
Omni-MATH threshold & 47.6874\\
Maximum rerolls / refractory windows & 3 / 2\\
Rollback margin & 64 tokens\\
Re-decode temperature / repetition penalty & 0.6 / 1.1\\
Suspect n-gram blocking & enabled\\
Prompt injection & disabled\\
\bottomrule
\end{tabular}
\end{table}

\section{Descriptive MATH-500 Results by Subject}

\begin{table}[H]
\centering
\caption{Three-seed MATH-500 accuracy by subject. These comparisons are descriptive and not multiplicity-adjusted.}
\small
\begin{tabular}{lrrr}
\toprule
Subject (units) & Vanilla & \method{} & Difference\\
\midrule
Algebra (372) & 68.01\% & 69.35\% & +1.34\pp\\
Counting and probability (114) & 42.98\% & 44.74\% & +1.75\pp\\
Geometry (123) & 47.15\% & 46.34\% & $-0.81$\pp\\
Intermediate algebra (291) & 42.61\% & 45.36\% & +2.75\pp\\
Number theory (186) & 62.37\% & 60.22\% & $-2.15$\pp\\
Prealgebra (246) & 55.69\% & 60.16\% & +4.47\pp\\
Precalculus (168) & 50.00\% & 52.38\% & +2.38\pp\\
\bottomrule
\end{tabular}
\end{table}

\section{Artifact Provenance}

MATH-500 uses dataset revision \hashid{6e4ed1a2a79af7d8630a6b768ec859cb5af4d3be} and manifest SHA-256 \hashid{b0302d2b75bcc3980f459abb260e8de8d1671d294b160acbeecfe2c2ea5a9baf}. The seed-0 and seeds-1--2 freeze hashes are \hashid{d26ba748646beab47be49c3270dc069640c9c649d1423fe54ab03426950f5482} and \hashid{500b0a9fb09038aa0ccd9fcf46b486fed2840e3c090004ff4893ada5cb490fb4}, respectively. The MATH-500 ablation environment used Python 3.10.20, PyTorch 2.13.0+cu130, Transformers 4.57.6, Datasets 5.0.1, bitsandbytes 0.50.1, and CUDA 13.0.

Omni-MATH uses source commit \hashid{23be225c8e268df51990f6c5c1448f34d3b56911}, test manifest \hashid{0d84b7670bf640257517e089c51ffac159a144ad1dccce11562a087a467a4c51}, final freeze \hashid{fce1576a35011fc6eb39c25839e997e742379837d6f1e193f217b55ee84d29d8}, and source-code commit \hashid{478c3ec49f65a0d38eadc6cf922b5fe8263df853}. Its environment used Python 3.9.25, PyTorch 2.8.0+cu128, Transformers 4.57.6, Datasets 4.5.0, bitsandbytes 0.48.2, and CUDA 12.8.


\begin{thebibliography}{99}

\bibitem[Anonymous(2026)]{anonymous2026prior}
Anonymous. Prior work on calibrated e-CUSUM decoding for quantized reasoning models. Anonymized preprint, 2026. Bibliographic details withheld for double-blind review.

\bibitem[DeepSeek-AI(2025)]{deepseek2025r1}
DeepSeek-AI. DeepSeek-R1: Incentivizing reasoning capability in LLMs via reinforcement learning. \emph{arXiv:2501.12948}, 2025.

\bibitem[Duan et~al.(2026)]{duan2026circular}
Zenghao Duan et~al. Circular reasoning: Understanding self-reinforcing loops in large reasoning models. \emph{arXiv:2601.05693}, 2026.

\bibitem[Guan et~al.(2026)]{guan2026deviation}
Weixin Guan et~al. Mitigating overthinking in large reasoning language models via reasoning path deviation monitoring. \emph{arXiv:2603.14251}, 2026.

\bibitem[Holtzman et~al.(2020)]{holtzman2020curious}
Ari Holtzman, Jan Buys, Li Du, Maxwell Forbes, and Yejin Choi. The curious case of neural text degeneration. \emph{ICLR}, 2020.

\bibitem[Li et~al.(2025)]{li2025quantization}
Zhen Li et~al. Quantization meets reasoning: Exploring and mitigating degradation of low-bit LLMs in mathematical reasoning. \emph{arXiv:2505.11574}, 2025.

\bibitem[Lightman et~al.(2024)]{lightman2024verify}
Hunter Lightman et~al. Let's verify step by step. \emph{ICLR}, 2024.

\bibitem[Liu et~al.(2025)]{liu2025quantization}
Ruikang Liu et~al. Quantization hurts reasoning? An empirical study on quantized reasoning models. \emph{arXiv:2504.04823}, 2025.

\bibitem[Lotfi et~al.(2026)]{lotfi2026quantized}
Sanae Lotfi, Polina Kirichenko, Steven Li, and Zechun Liu. Quantized reasoning models think they need to think longer, but they do not. \emph{arXiv:2606.00206}, 2026.

\bibitem[Madaan et~al.(2023)]{madaan2023selfrefine}
Aman Madaan et~al. Self-Refine: Iterative refinement with self-feedback. \emph{NeurIPS}, 2023.

\bibitem[Min et~al.(2026)]{min2026puma}
Donghyeon Min et~al. Stop when reasoning converges: Semantic-preserving early exit for reasoning models. \emph{arXiv:2605.17672}, 2026.

\bibitem[Page(1954)]{page1954continuous}
E. S. Page. Continuous inspection schemes. \emph{Biometrika}, 41(1/2):100--115, 1954.

\bibitem[Ramdas et~al.(2023)]{ramdas2023gametheoretic}
Aaditya Ramdas, Peter Gr\"unwald, Vladimir Vovk, and Glenn Shafer. Game-theoretic statistics and safe anytime-valid inference. \emph{Statistical Science}, 38(4):576--601, 2023.

\bibitem[Rezazadeh and Davoodi(2026)]{rezazadeh2026shape}
Nima Rezazadeh and Ali G. Davoodi. The shape of overthinking: Backtracking bursts in long reasoning traces. \emph{arXiv:2605.27965}, 2026.

\bibitem[Shafer(2021)]{shafer2021testing}
Glenn Shafer. Testing by betting: A strategy for statistical and scientific communication. \emph{JRSS A}, 184(2):407--431, 2021.

\bibitem[Shin et~al.(2024)]{shin2024edetectors}
Jaehyeok Shin, Aaditya Ramdas, and Alessandro Rinaldo. E-detectors: A nonparametric framework for sequential change detection. \emph{NEJSDS}, 2:229--260, 2024.

\bibitem[Snell et~al.(2024)]{snell2024scaling}
Charlie Snell, Jaehoon Lee, Kelvin Xu, and Aviral Kumar. Scaling LLM test-time compute optimally can be more effective than scaling model parameters. \emph{arXiv:2408.03314}, 2024.

\bibitem[Tran et~al.(2026)]{tran2026frontier}
D. Tran et~al. Beyond the frontier: Stochastic backtracking for efficient test-time scaling. \emph{arXiv:2605.25143}, 2026.

\bibitem[Wang et~al.(2023)]{wang2023selfconsistency}
Xuezhi Wang et~al. Self-consistency improves chain of thought reasoning in language models. \emph{ICLR}, 2023.

\bibitem[Wei et~al.(2026)]{wei2026evolution}
Zihao Wei et~al. The evolution of thought: Tracking LLM overthinking via reasoning dynamics analysis. \emph{ACL}, 2026.

\bibitem[Yang et~al.(2025)]{yang2025selfbacktracking}
Xiao-Wen Yang et~al. Step back to leap forward: Self-backtracking for boosting reasoning of language models. \emph{arXiv:2502.04404}, 2025.

\bibitem[Yao et~al.(2023)]{yao2023tree}
Shunyu Yao et~al. Tree of thoughts: Deliberate problem solving with large language models. \emph{NeurIPS}, 2023.

\end{thebibliography}
\end{document}